\documentclass{article}
\usepackage{spconf,amsmath,graphicx,amssymb,bm}
\usepackage{booktabs}
\usepackage{multirow}
\usepackage[ruled]{algorithm2e}
\usepackage{url}
\usepackage{caption}
\usepackage{subcaption}

\usepackage{threeparttable}
\usepackage{enumitem}
\usepackage{verbatim}

\def\vk{{\bm{k}}}

\def\vq{{\bm{q}}}

\def\vu{{\bm{u}}}
\def\vv{{\bm{v}}}
\def\vw{{\bm{w}}}
\def\vx{{\bm{x}}}
\def\vy{{\bm{y}}}

\def\mA{{\bm{A}}}

\def\mK{{\bm{K}}}

\def\mQ{{\bm{Q}}}

\def\mV{{\bm{V}}}
\def\mW{{\bm{W}}}
\def\mX{{\bm{X}}}

\def \RR {{\mathbb{R}}}

\newtheorem{theorem}{Theorem}

\newtheorem{lemma}{Lemma}

\usepackage{color}

\title{GlassoFormer: a Query-Sparse 
Transformer for\\ 
Post-Fault Power Grid Voltage Prediction}
\name{Yunling Zheng$^1$, Carson Hu$^1$, Guang Lin$^2$, Meng Yue$^3$, Bao Wang$^4$, Jack Xin$^1$
\thanks{The work was partially supported by NSF and DOE grants DMS-1924935, DMS-1952644, DMS-1924548,  DMS-1952339, 
DE-SC0021142.}
\thanks{
Accepted by IEEE ICASSP 2022}}
\address{
\normalsize $^1$ Department of Mathematics, University of California Irvine\\
\normalsize $^2$ Department of Mathematics and School of Mechanical Engineering, Purdue University\\
\normalsize $^3$ Interdisciplinary Science Department, Brookhaven National Laboratory\\
\normalsize $^4$ Department of Mathematics and Scientific Computing and Imaging Institute, The University of Utah
}
\begin{document}

\setlength{\abovedisplayskip}{3pt}
\setlength{\belowdisplayskip}{3pt}
%
\maketitle
\begin{abstract}
We propose GLassoformer, a novel and efficient transformer architecture leveraging group Lasso regularization to reduce the number of queries of the standard self-attention mechanism. Due to the sparsified queries, GLassoformer is more computationally efficient than the standard transformers. On the power grid post-fault voltage prediction task, GLassoformer shows remarkably better prediction than many existing benchmark algorithms in terms of accuracy and stability.

\end{abstract}
\begin{keywords}
efficient transformer, query sparsity, group lasso, power grid prediction
\end{keywords}
\section{Introduction}
\label{sec:intro}

\begin{figure}[ht]
	\vspace{-0.6cm}
	\centering
	\includegraphics[width=0.6\linewidth]{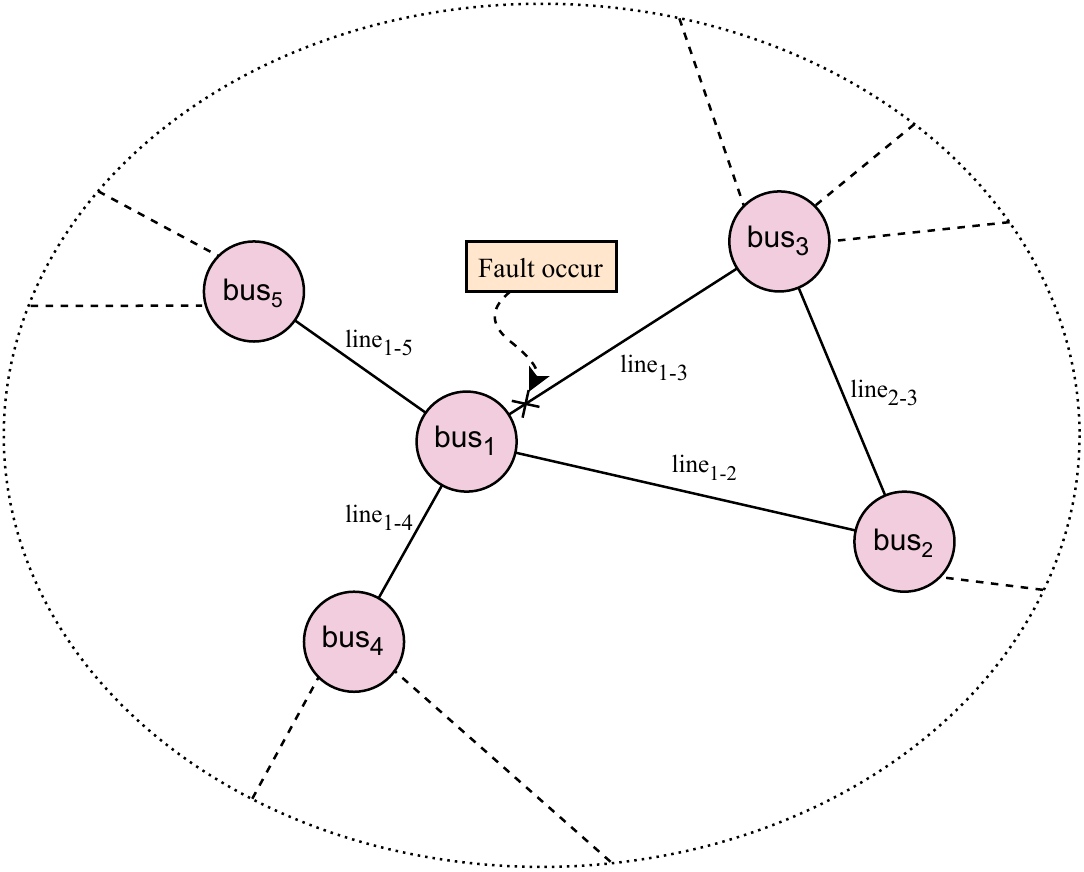}
	\vspace{-0.3cm}
	\caption{\small Snapshot illustrates a faulty power grid system with voltages/currents recorded by sensors on buses/lines.}
	\vspace{-0.3cm}
	\label{fig:system}
\end{figure}

Lifeline networks such as power grids, transportation, and water networks are critical to normal functions of society and economics. However, rare events and disruptions occur due to natural disasters (earthquakes and hurricanes), aging and short circuits, to name a few. Knowledge of power grid transient stability after fault occurrences is crucial for decision making; see Fig.\ref{fig:system} for an illustration of a power grid system with a line fault. In particular, an online tool for predicting the transient dynamics, if available, will be very beneficial to the operation of the increasingly dynamic power grid due to the increasing un-dispatchable renewables. Computationally, a traditional approach for transient assessment is to simulate a large physical system of circuit equations. The simulation-based approach is very time-consuming and requires detailed fault information such as type, location, and clearing times of fault. Therefore, it is not applicable for online assessment due to the unavailability of the fault information. 

With the increasing deployment of phasor measurement units (PMUs) in power grids, high-resolution measurements provide an alternative data-driven solution, e.g., learning the post-fault system responses and predicting subsequent system trajectories based on post-fault short time observations. 
Machine learning (ML)-based online transient assessment methods have been developed using real-time measurements of the initial time-series responses as input. These ML-based studies aim to predict system stability by deriving a binary stability indicator \cite{wang2016,james2017,yan2019} or estimating the stability margin \cite{liu2013,lotufo2007,zhu2019}. However, knowing the system stability or the margin only is often not enough. Instead, having the knowledge of post-fault transient trajectories is more important for the system operators \cite{MLGDyn} to take appropriate actions, e.g., a load shedding upon a frequency or voltage violation.

Classical signal processing relies on linear time-invariant filters, such as Pad\'e and Prony's methods \cite{Hayes}, and can be used for online prediction. However, these methods are limited to fitting responses with a rational function, requiring users to choose integer parameters (filter orders) from observed data. The prediction is not robust with respect to such choices. In \cite{pwgrd_1dcnn}, Prony's method is generalized to 1D convolutional neural networks (1D-CNN) with additional spatial coupling (e.g., currents in power lines nearby). Temporal (1D) convolution exists in both methods, yet 1D-CNN also contains nonlinear operations and more depths in feature extractions. Predictions by 1D-CNN improve over Prony's method considerably \cite{pwgrd_1dcnn}. To go further along this line, we notice that transformers (\cite{vaswani2017attention,Context_Sparse_Tran_20} and references therein) have achieved state-of-the-art performance in machine translation and language modeling due to their more non-local representation power than convolutions. However, transformers suffer from quadratic costs in computational time and memory footprint with respect to the sequence length, which can be prohibitively expensive when learning long sequences. Moreover, redundant queries in transformers can degrade the prediction accuracy. 
Leveraging group Lasso-based structured sparsity regularization, we optimize the sparsity of queries of the self-attention mechanism, resulting in a reduced number of queries and computational cost. We further integrate 1D-CNN layers into transformers with sparse queries for power grid prediction and show remarkable improvement over existing methods.

\vspace{-0.2cm}
\section{Methodology}
\label{sec:meth}

The proposed model is based on the encoder-decoder architecture \cite{vaswani2017attention} with efficient attention. An overview is shown in the top panel of Fig. 2, and more details are described below.

\vspace{-0.2cm}
\subsection{Transformer}
\vspace{-0.1cm}
{\bf Embedding.}
In the voltage prediction of the post-fault power grid system, the ability to capture long-term voltage volatility patterns requires both global information like time stamps and early-stage temporal time-series features from connected neighbors. To this end, we use a uniform representation embedding layer based on padding and convolution to mitigate mismatch between time stamps and temporal inputs in the encoder and decoder.

The input signals of a single bus in a fault event is $\mX=\{\dots,\vx_i,\dots\},\ 1 \leq i \leq F$ and $ \vx_i \in \mathbb{R}^{L}$. Where $F$ is he number of input signals, counting both voltage and current connected or adjacent to the bus and $L$ is length of signal measured. The global time stamp $\vx^{s} \in \mathbb{R}^{L}$ records temporal (positional) relationship to the fault time $t_f$.
The encoder input is $\vx^{en}=\{\vx; \vx^{s}\}$. For input of decoder, only the features of initial period before time $t_f$ is used. We take zero padding to match dimension, and get $ \vx^{de} = \{\dots, \vx_i^{de}, \dots; \vx^s\} $, where $ \vx_i^{de} = \{x_{i, t_0}, \dots, x_{i, t_f}, 0, \dots, 0\} $.
This embedding ensures causality and is written as
\[ \mX = \text{ELU}(\text{Concat}(\text{Conv1d}(\vx_{in}), \text{Conv1d}(\vx^{s}))), \]
where $\vx_{in}$ is either $\vx^{en}$ or $\vx^{de}$; and ELU$(u)=\exp\{u\} -1 $ if $u< 0$, ELU$(u)=u$ if $u\geq 0$,  component-wise on a vector.

\begin{figure}[!ht]
\vspace{-0.3cm}
\centering
\begin{subfigure}{0.8\columnwidth}
	\centering
	\includegraphics[width=\textwidth]{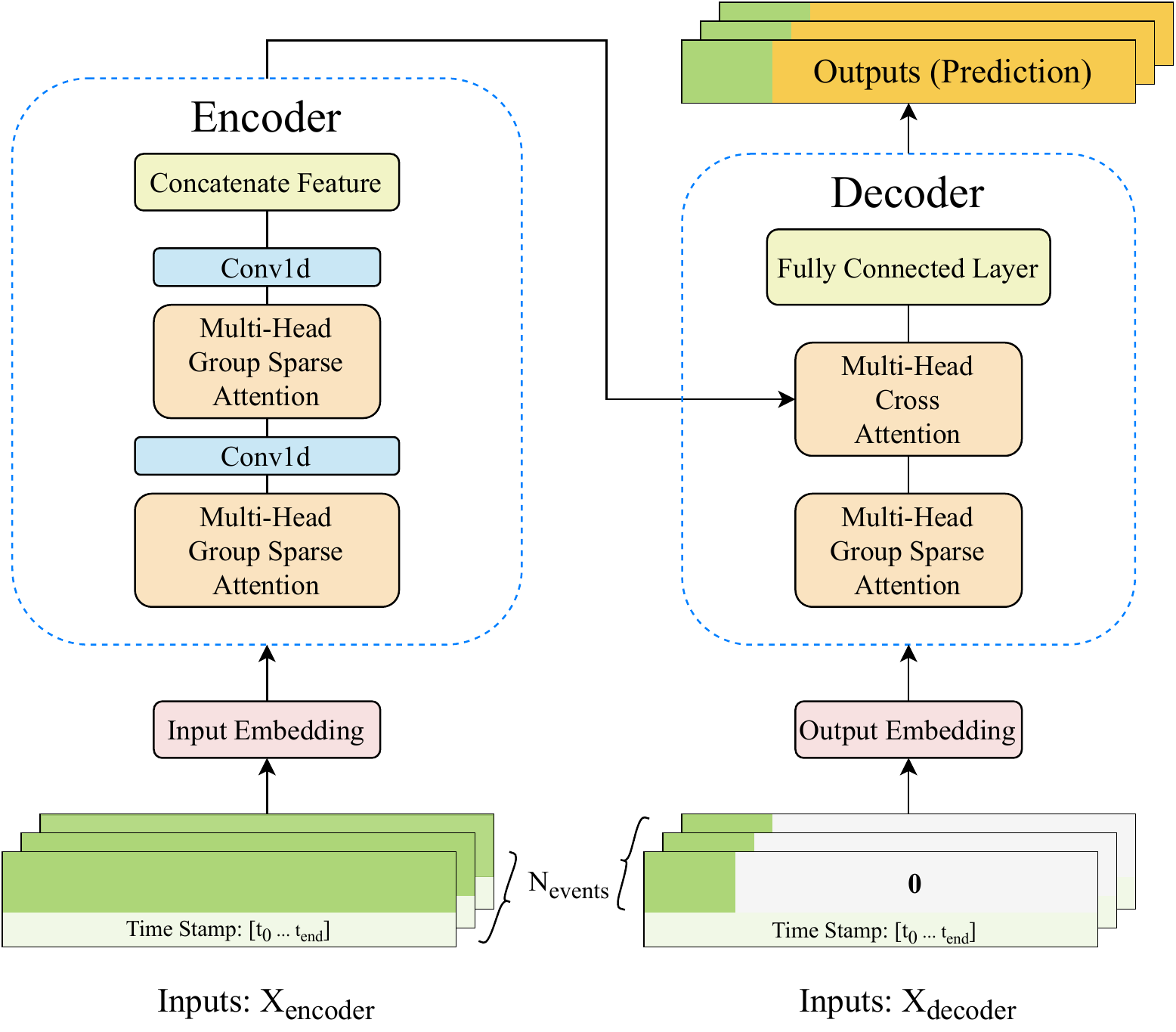}	
\end{subfigure}
\begin{subfigure}{0.8\columnwidth}
	\vspace{0.2cm}
	\centering
	\includegraphics[width=\textwidth]{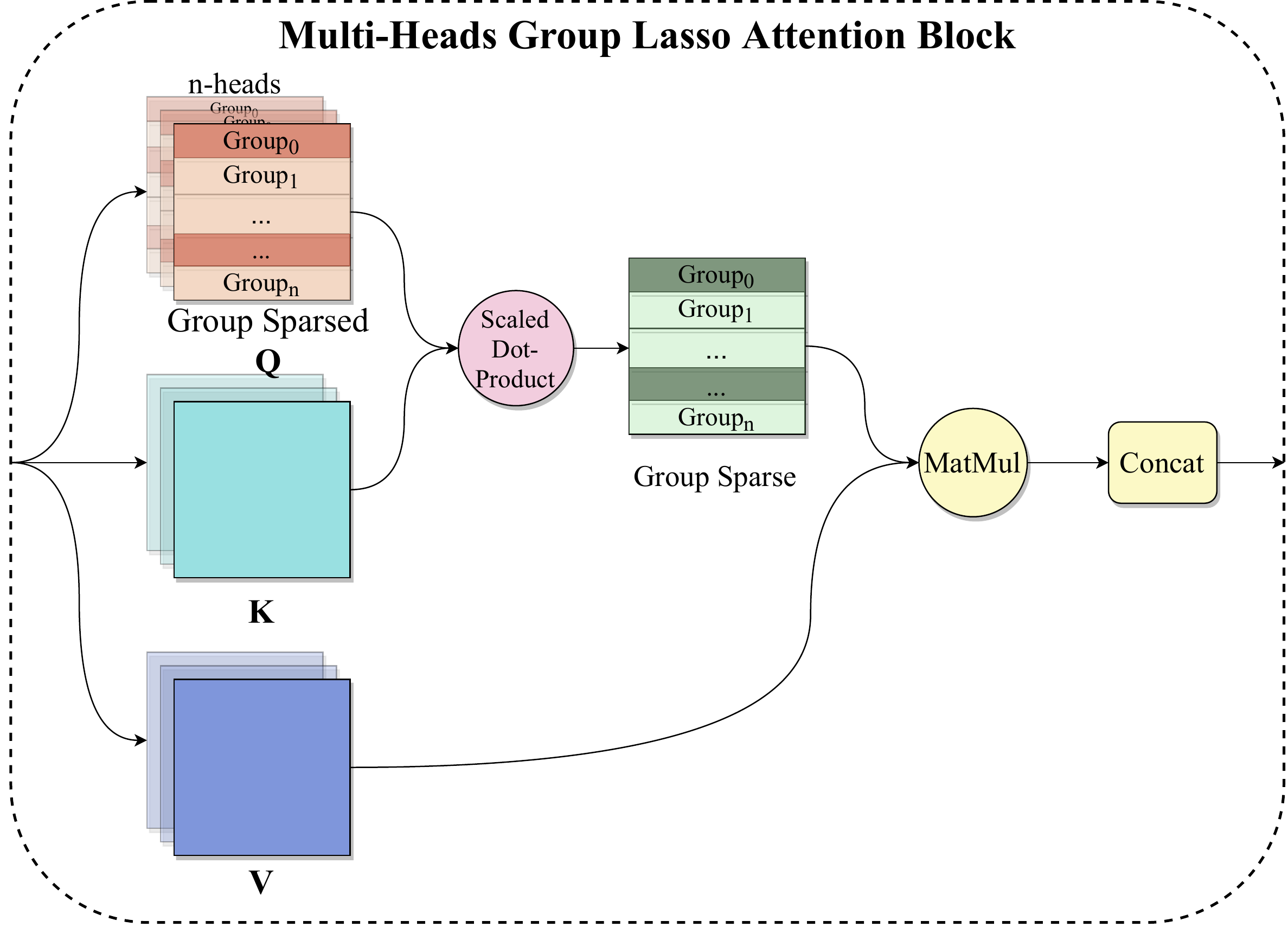}
\end{subfigure}
\vspace{-0.2cm}
\caption{\small Top panel: overall encoder-decoder architecture of GLassoformer, bottom panel:  GLassoformer's attention block.}
\label{fig:model}
\end{figure}

\medskip
\noindent{\bf Encoder.}
After embedding and representations, the input signal is formed as $ \mX_{en} \in \mathbb{R}^{L\times d_{model}} $, see
Fig.\ref{fig:model} (top).
In the encoder, there are two identical layers; each consists of multi-head Group Sparse attention and a 1-D convolution layer. The sub-layer, Group Sparse attention, is discussed in detail in Sec.\ref{sec:sparseatten} for removing redundancy in the vanilla attention module \cite{vaswani2017attention}. The 2nd sub-layer going forward is written as
\[ \text{ELU}(\text{Conv1d}(\text{Attn}(X_{en}))) \]
where Conv1d performs 1-D convolutional  filtering in time dimension \cite{Hayes}, followed by the ELU activation function \cite{Informer_21}. 

\medskip
\noindent{\bf Decoder.}
The embedding layer before decoder shapes the inputs feature as $ \mX_{de} \in \mathbb{R}^{L\times d_{model}} $, where $d_{model}$ is the hidden dimension of model. This layer enables Group Sparse attention to perform the same as in the encoder. After multi-head Group Sparse attention, a canonical multi-head cross attention \cite{Informer_21} combines the hidden feature from encoder and $\text{Attn}(\mX_{de})$. 
A fully connected layer after cross attention matches the output dimension with that of the prediction variables. The final output is
\[ \vy^n = \text{Full}(\text{CrossAttn}(F_{hidden}, \text{Attn}(\mX_{de}))), \]
where Full and CrossAttn stand for the fully connected and cross attention layers respectively, and $ F_{hidden} $ is the hidden feature from the  encoder layer.


\vspace{-0.2cm}
\subsection{Self-attention mechanism and query sparsification}
\vspace{-0.1cm}
\label{sec:sparseatten}
The self-attention mechanism \cite{vaswani2017attention} is used to learn long-range dependencies and enable parallel processing of the input sequence. For a given input sequence $\tilde{\mX}:=[\tilde{\vx}_1,\cdots,\tilde{\vx}_N]^\top,\ \tilde{\vx} \in \RR^{D_x}$, 
self-attention transforms $\tilde{\mX}$ into an output sequence $\Hat{\mV}$ in the following two steps\footnote{For simplicity of the following discussion, we formulate the self-attention mechanism slightly differently from that in \cite{vaswani2017attention}.}:

\vspace{-0.2cm}
\begin{enumerate}
    \item [Step 1.] Project the input sequence $\tilde{\mX}$ into three matrices via the following linear transformations 
    $$
\mQ=\mW_Q^\top\, \tilde{\mX}^\top; \mK=\mW_K^\top\, \tilde{\mX}^\top; 
\mV=\mW_V^\top\, \tilde{\mX}^\top,
$$
where $\mW_Q,\mW_K, \mW_V \in \RR^{D_x \times N}$ are the weight matrices. We denote $N\times N$ matrices $\mQ:=[\vq_1,\cdots,\vq_N]^\top$, $\mK:=[\vk_1,\cdots,\vk_N]^\top$, and $\mV:=[\vv_1,\cdots,\vv_N]^\top$, where the vectors $\vq_i,\vk_i,\vv_i$ for $i=1,\cdots,N$ are query, key, and value vectors respectively.
    
    \item [Step 2.] For each query vector $\vq_i$ for $i=1,\cdots,N$, we compute the output vector $\Hat{\vv}_i$ as follows
\vspace{-0.3cm}
\begin{equation}\label{eq:attention-vec}
\hat{\vv}_i=\sum_{j=1}^N{\rm softmax}\Big(\frac{{\vq}_i^\top{\vk}_j}{\sqrt{D}}\Big){\vv}_j,
\end{equation}
\vspace{-0.3cm}
i.e.,
\vspace{-0.1cm}
\begin{equation}
	\label{eq:attention-mat}
	\hat{\mV}={\rm softmax}\Big(\frac{{\mQ}{\mK}^\top }{\sqrt{D}}\Big){\bf V} :={\mA}{\mV},
\end{equation}
where the softmax 
is applied row-wise.
\end{enumerate}

\vspace{-0.3cm}
For long sequences, the computational and memory costs of transformers are dominated by \eqref{eq:attention-vec}. It is evident that the memory cost is $\mathcal{O}(N^2)$ to store the attention matrix $\mA$. Also, the computational complexities of computing the matrix-matrix products $\mQ\mK^\top$ and $\mA\mV$ are both $\mathcal{O}(N^2)$. In response, efficient transformers have been proposed leveraging low-rank and/or sparse approximation of 
$\mA$ \cite{katharopoulos2020transformers,performer,beltagy2020longformer,ainslie-etal-2020-etc}, locality-sensitive hashing \cite{Kitaev2020Reformer}, clustered attention \cite{vyas2020fast}, etc. 

 In \cite{Context_Sparse_Tran_20},  $k$-means clustering identifies most relevant keys to reduce the number of query-key pairs.
In \cite{Informer_21}, an empirical formula based on Kullback-Leibler divergence is derived to score the likeness of $\mQ$ and $\mK$, 
and select top few row vectors in 
$\mQ$. To reduce the score computation to $\mathcal{O}(N \log N )$ 
complexity, random sampling of $\mQ$ vectors is used to approximate the score formula. The resulting top score $\mQ$ vectors remain and others are zeroed out. Such a group sparsification of $\mQ$ helps remove insignificant part of the full attention matrix, also lower computation and storage costs to $\mathcal{O}(N \log N)$. 

We observe from \eqref{eq:attention-mat} that if query matrix $\mQ$ only has $\mathcal{O}(1)$ many nonzero rows (or group sparsity), the complexity of $\mA\mV$ goes down to $\mathcal{O}(N)$.
This amounts to replacing the empirical score formula in \cite{Informer_21} by a data-driven row selection in $\mQ$. To this end, we adopt the group Lasso (GLasso \cite{GL_2007}) penalty {\it to sparsify columns of
$\mW_{Q}$ so that the number of nonzero rows of $\mQ = \mW_{Q}^\top \tilde{\mX}^\top$ is reduced to $\mathcal{O}(1)$ which in turn lowers complexity of $\mA\, \mV$ to $\mathcal{O}(N)$.} 
Though GLasso computation without any query vector sampling does not reduce 
$\mathcal{O}(N^2)$ complexity of 
attention computation in training (it surely does so at inference), this is a minor problem for our moderate-size data set and network with a few million parameters in this study. The training times are about the same with or without a random sampling of query vectors on the power grid data set (Tab. \ref{tab:sprasetime}). 
More importantly, our resulting network (GLassoformer) is faster and more accurate 
than Informer \cite{Informer_21} during testing and inference, see Tabs. \ref{tab:result} and
\ref{tab:paramcompare}. 
%
%
%
%
\section{Algorithm and Convergence}

We present our algorithm to realize row-wise sparsity in query matrix 
$\mQ$ and show its convergence guarantee.
\subsection{Algorithm}
\vspace{-0.1cm}
Let $ (\theta, \mW_Q) $ denote model parameters, where $ \mW_Q$ is the query weights in the attention mechanism. 
We divide $ \mW_Q $ into column-wise groups: $ \mW_Q = \{\vw_1,\dots,\vw_g,\dots,\vw_N\} $. %
The penalized total loss is:
\begin{equation}\label{ploss}
	L(\theta, \mW_Q) = f(\theta, \mW_Q) + \lambda \, \left\lVert \mW_Q\right\rVert _{GL}
\end{equation}
where $f$ is 
the 
mean squared 
loss function, and  
$ \left\lVert \mW_Q \right\rVert_{GL}  \triangleq \sum_{g=1}^{N} \left\lVert \vw_g\right\rVert _{2}  $ is the GLasso penalty \cite{GL_2007}.
For network training, we employ the Relaxed Group-wise Splitting Method (RGSM, \cite{rgsm}) which outperforms standard gradient descent on related group sparsity (channel pruning) task of convolutional networks for loss functions of the type (\ref{ploss}). To this end, we pose the proximal problem:
\begin{equation}\label{eq:glprox}
	\vy_{g}^{*} = \arg\min_{y_g} \lambda \left\lVert \vy_g\right\rVert _2 + \sum_{i\in I_g} \frac{1}{2} \left\lVert \vy_{g,i}-\vw_{g,i} \right\rVert _{2}^{2}
\end{equation}
 where $ I_g $ is the column index set of $ \vw $ in group $g$. Solution of (\ref{eq:glprox}) is a soft-thresholding function:
\begin{equation*}
	y_{g}^{*} = \mathbf{Prox}_{GL,\lambda} (\vw_g) \triangleq \vw_g \max(\left\lVert \vw_g\right\rVert _{2} - \lambda, \mathbf{0}) / \left\lVert \vw_g\right\rVert _2. 
\end{equation*}
The RGSM algorithm minimizes a relaxation of (\ref{ploss}):
\vspace{-0.3cm}
\begin{align} \label{relaxL}
L_\beta (\theta,\vw,\vu) := f(\theta, \vw) + \lambda \, \left\lVert \vu \right\rVert _{GL}
    +{\frac{\beta}{2}} \, \left\lVert \vw -\vu
    \right\rVert_{2}^{2}
    \vspace{-0.cm}
\end{align}
\vspace{-0.3cm}
by iterations:
\begin{align}
	\vu_{g}^{t} &= \mathbf{Prox}_{\lambda} (\vw_{g}^{t}), \text{\qquad for}\ g = 1,\dots,N \label{eq:descent} \\
	(\theta,\vw)^{t+1} &= (\theta,\vw)^{t} - \eta \, \nabla f(\theta^t, \vw^t) - (0, \, \eta\, \beta (\vw^t-\vu^t)) \nonumber
\end{align}
with $\nabla := \nabla_{\theta,\vw}$, learning rate $\eta $, relaxation parameter $\beta > 0$.

\vspace{-0.3cm}
\subsection{Convergence Theory}
Thanks to the ELU activation function 
(continuously differentiable with piece-wise bounded second derivative), 
the network mean squared loss function 
$f$ satisfies Lipschitz 
gradient inequality for some positive constant $L_{ip}$:
\begin{align}\label{Lip}
\| \nabla f(\vv_1) - 
\nabla  f(\vv_2) \| 
\leq L_{ip}\, 
\|\vv_1 - \vv_2\|
\end{align}
and any $\vv_i:=(\theta_i,\omega_i)$, $i=1,2$. Notice that $\|\cdot\|_{GL}$ violates (\ref{Lip}). The advantage of splitting in RGSM (\ref{eq:descent}) is to overcome this lack of non-smoothness in convergence theory. 
We state the following (the proof follows from applying  
(\ref{Lip}) and (\ref{eq:descent}), similar to Appendix A of \cite{sprob_20})
\begin{lemma}[Descent Inequality]
\begin{align*}
{    L_\beta( \vv^{t+1},\vu^t ) \leq L_\beta (\vv^t, \vu^t) + 
    \left(\frac{L_{ip}}{2}+\frac{\beta}{2}-\frac{1}{\eta}\right)\|\vv^{t+1}-\vv^t\|^2}
\end{align*}
\end{lemma}
\vspace{-0.2cm}
which implies
\vspace{-0.1cm}
\begin{theorem}
If $f$ is coercive ($f$ bounded implies that of its independent variables, true when standard weight decay is present in network training) and the learning rate $\eta < {2}/{\beta+L_{ip}}$, then $L_{\beta}(\vv^t,\vu^t)$ decreases monotonically in $t$, and 
$(\vv^t,\vu^t)$ converges sub-sequentially to a limit point $(\bar{\vv},\bar{\vu})$, from which $\bar{\vw}$ is extracted to speed up inference.
\end{theorem}

\section{Experimental Settings}
\label{sec:expe}

{\bf Datasets.}\ 
We carry out experiments 
 on the 
simulated New York/New England 16-generator 68-bus power system 
\cite{PST}.
The data set takes over 2248 fault events, where each event has signals of 10 seconds long. These signals contain voltage and frequency from every bus, current from every line.
The dataset also records the graph structure of buses and lines, see Fig.~\ref{fig:system}.
The system has 68 buses and 88 lines linking them, i.e., 68 nodes with 88 edges in the graph. To explore the prediction accuracy of the GLassoformer model, we take the voltage of a bus as the training target, and voltage with currents from locally connected buses and lines as input features. The splitting for train/val/test is 1000/350/750 fault events.
\medskip


\noindent{\bf Experimental Details.}
We implemented our model on PyTorch. 
\textbf{Baselines:} we selected several time-series  prediction models to compare, including 1D-CNN, Informer, Lasso\footnote{Transformer with regular (un-structured) Lasso penalty.}, and Prony's method (for voltage signal of single bus, case II in Tab~\ref{tab:result}). 
\textbf{Architecture:} 
The encoder contains 2 Group Sparse attention layers, and decoder consists of 1-layer Group Sparse self-attention and 1-layer cross-attention. 
We use Adam with a learning rate $ \eta $ of $1e-4$ and rate decay by $0.8$ after every 10 epochs. 
For hyper-parameters in group Lasso, $ \beta $ is 0.9 and $ \lambda $ is $0.01$.
During training, we use a batch size of 30. The maximal training epoch number is 80 with early stopping patience of 30. 
\textbf{Setup:} The input is zero-mean normalized. 
\textbf{Platform:} The model is trained on Nvidia GTX-1080Ti. 




\vspace{-0.2cm}
\section{Results}
\begin{table}\fontsize{8.7}{8.7}\selectfont
	\centering
	\begin{tabular}{c@{\hspace{0.15cm}}c|c@{\hspace{0.15cm}}c@{\hspace{0.15cm}}c@{\hspace{0.15cm}}c@{\hspace{0.15cm}}c}
		\toprule
		\multicolumn{2}{c}{Models} & GLasso
		& Informer & Lasso
		& 1DCNN & Prony  \\
		\midrule
		\multirow{2}{*}{I} 
		& MSE($\times 10^{-5}$) &\textbf{3.189} &3.662 &3.532 &8.014 &  -- \\
		& MAE($\times 10^{-3}$) &\textbf{2.374} &2.684 &2.543 &6.087 & -- \\
		\midrule
		\multirow{2}{*}{II} 
		& MSE($\times 10^{-5}$) &\textbf{6.115} &6.599 &6.501 &17.21 & 397.6 \\
		& MAE($\times 10^{-3}$) &\textbf{3.520} &3.877 &3.611 &9.264 & 47.3 \\
		\bottomrule
	\end{tabular}\vspace{-0.3cm}
	\caption{\small Voltage prediction error comparison. 
	}
	\footnotesize{I (II): input data with (without) neighbor voltage and current features. 
	}
	\vspace{-0.3cm}
	\label{tab:result}
\end{table}

Fig.~4 shows that group and regular Lasso affect our network training as seen from visualizations of structured/unstructured sparse query $\mQ$ (left/right), where entries with magnitudes below  1e-5 are zeroed out. In Fig.~3, two sample predictions from GLassoformer  (green) and Informer (orange) are compared with the blue test data 
(to the right of the vertical dashed line). The blue curve to the left of the dashed line is short time observation after the fault (network input). Informer's prediction contains several spurious spikes that conceivably come from the random sampling procedure in their attention module for linear complexity.  In Tab. 3, we see that the Informer's training time is slightly shorter even though it has three attention layers, while G/LassoFormer has two attention layers in the encoder. The training time saving is limited on our data set.  However, GLassoformer has much better prediction accuracies with and without nearby voltage and line current features, as shown in Tab. 1 in terms of mean squares error (MSE) and mean absolute error (MAE). The Glassformer model is smaller in parameter size and faster at inference than Informer (Tab. 2).



\begin{figure}[!ht]
	\vspace{-0.5cm}
	\centering
	\begin{tabular}{c@{\hspace{0.3cm}}c}
	\hspace{-0.4cm}\includegraphics[width=0.54\columnwidth]{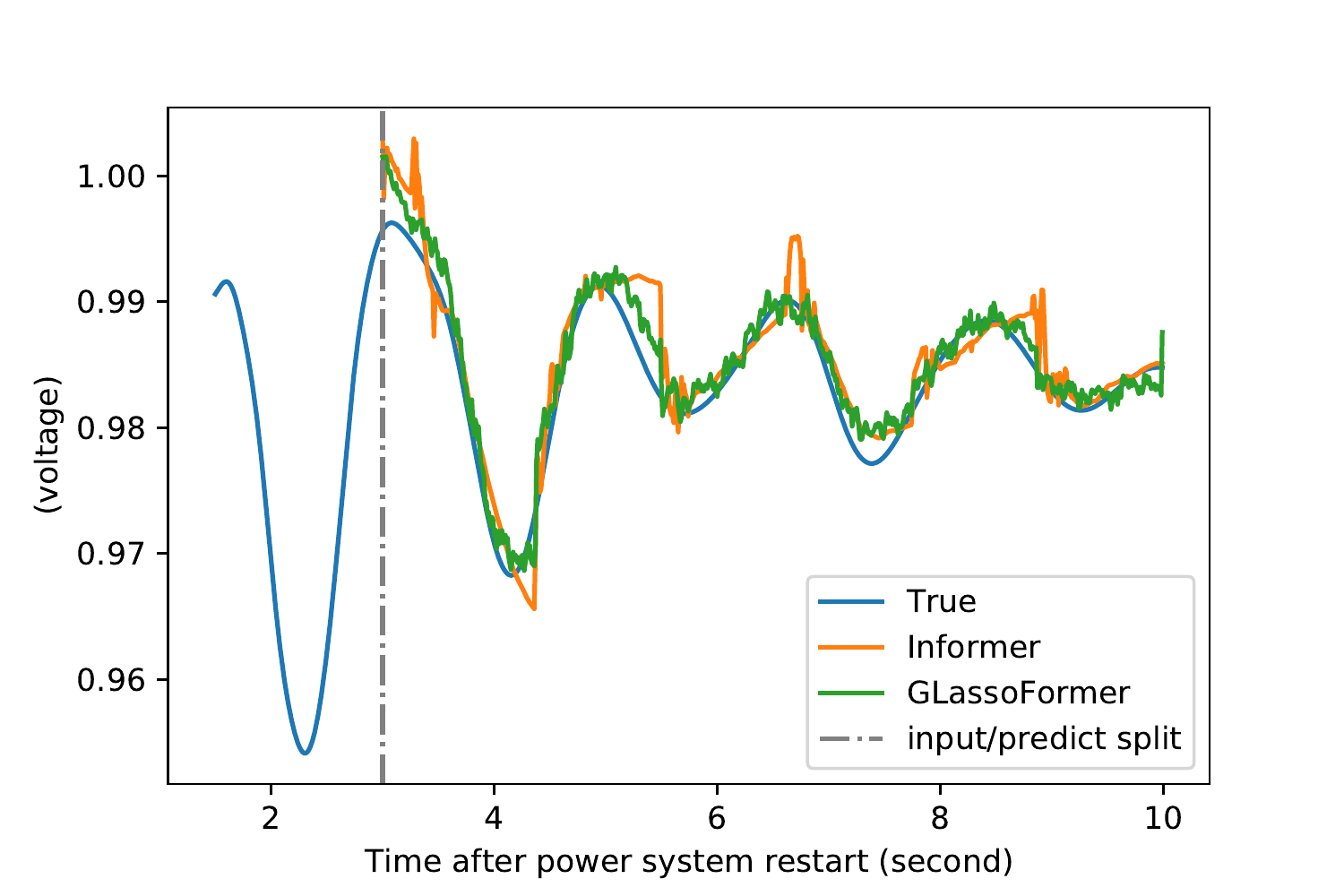}&
	\hspace{-0.7cm}\includegraphics[width=0.54\columnwidth]{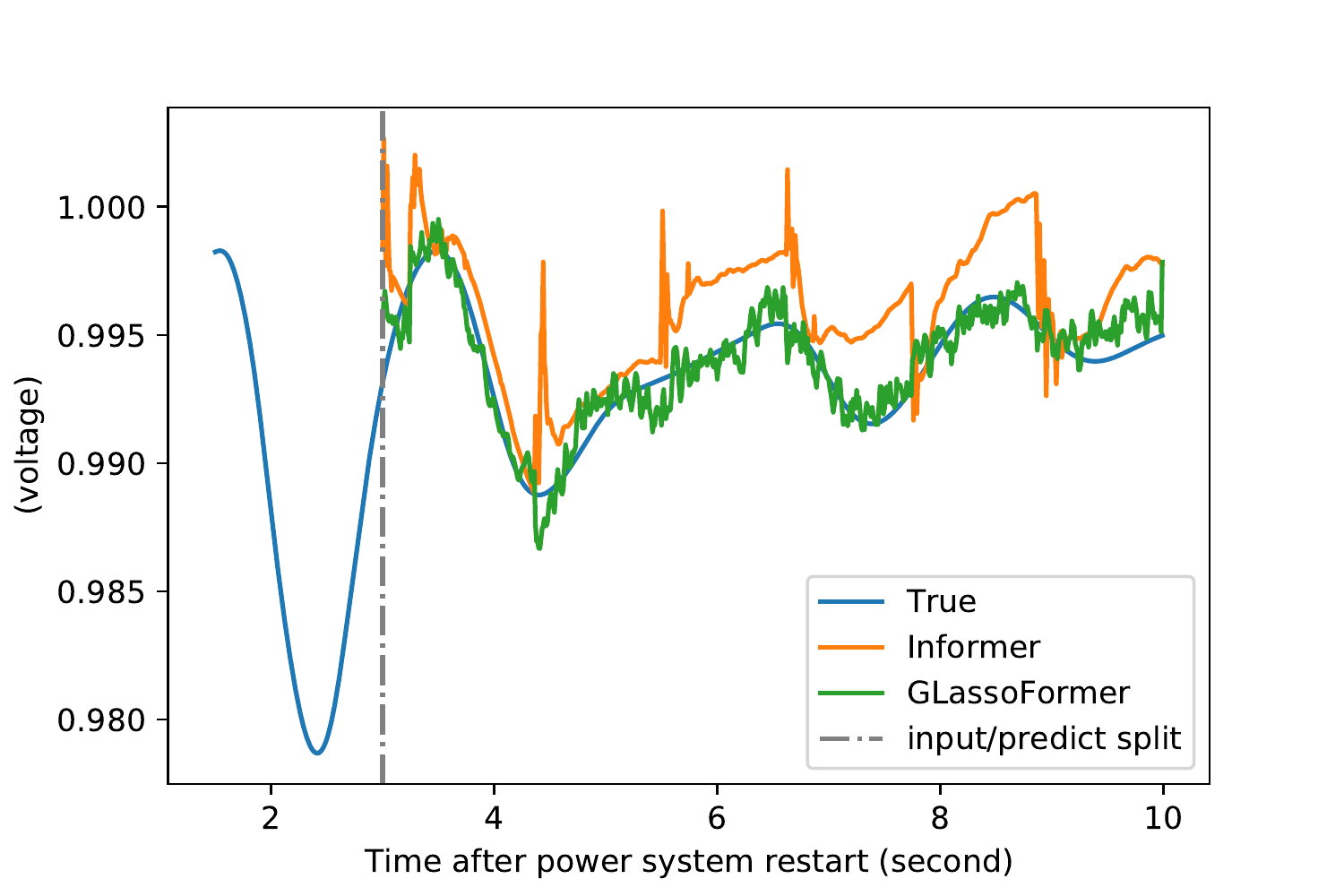}\\
	\end{tabular}
    \vspace{-0.5cm}
	\caption{\small Sample post-fault voltage predictions in time (to the right of the vertical dashed line at time 3) with input from the left of the line.}
\label{fig:comparison}
\end{figure}

\begin{figure}[!ht]
	\vspace{-0.9cm}
	\centering
	\begin{tabular}{@{\hspace{-0.35cm}}c@{\hspace{-0.65cm}}c}
	\includegraphics[width=0.6\columnwidth]{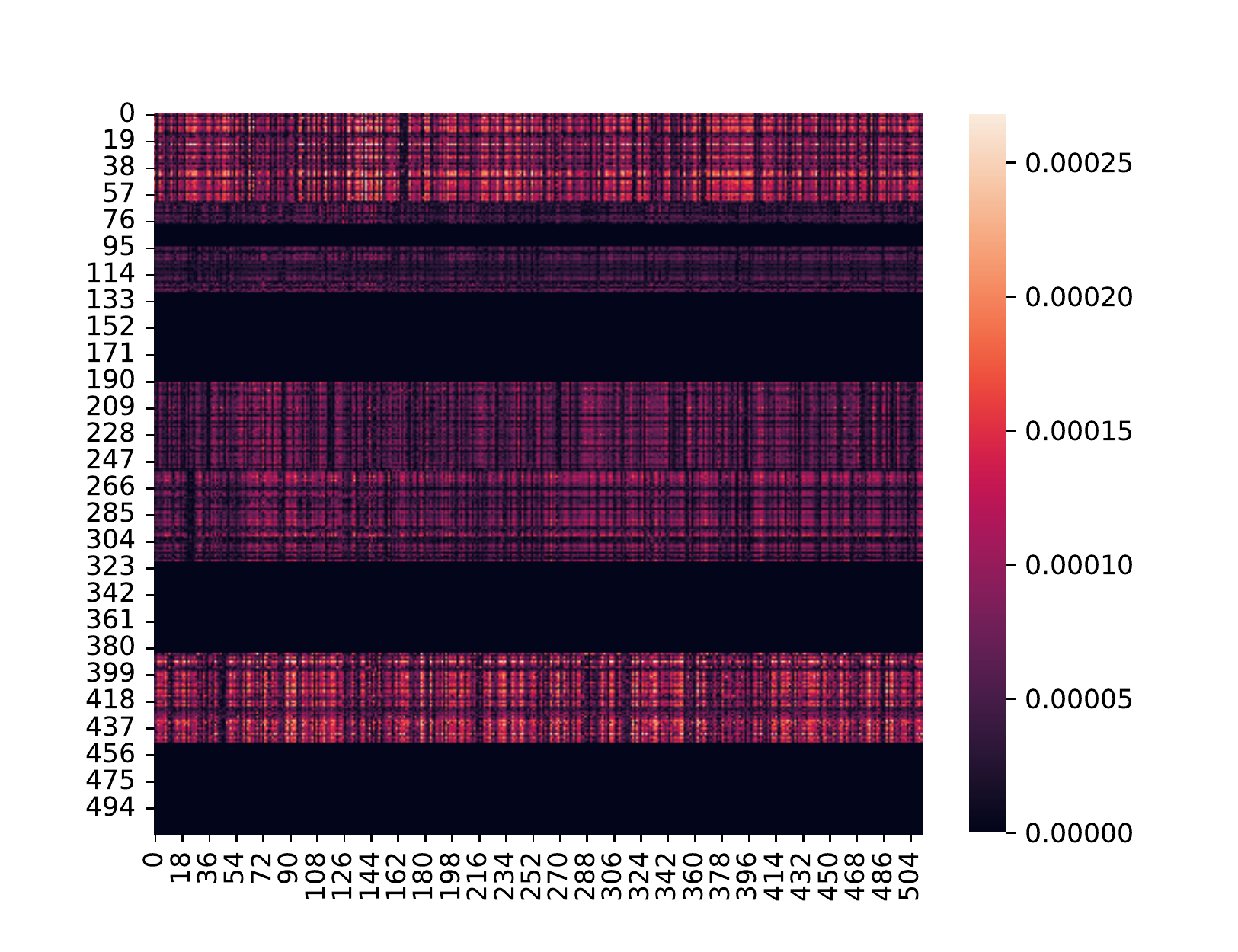}&
	\includegraphics[width=0.6\columnwidth]{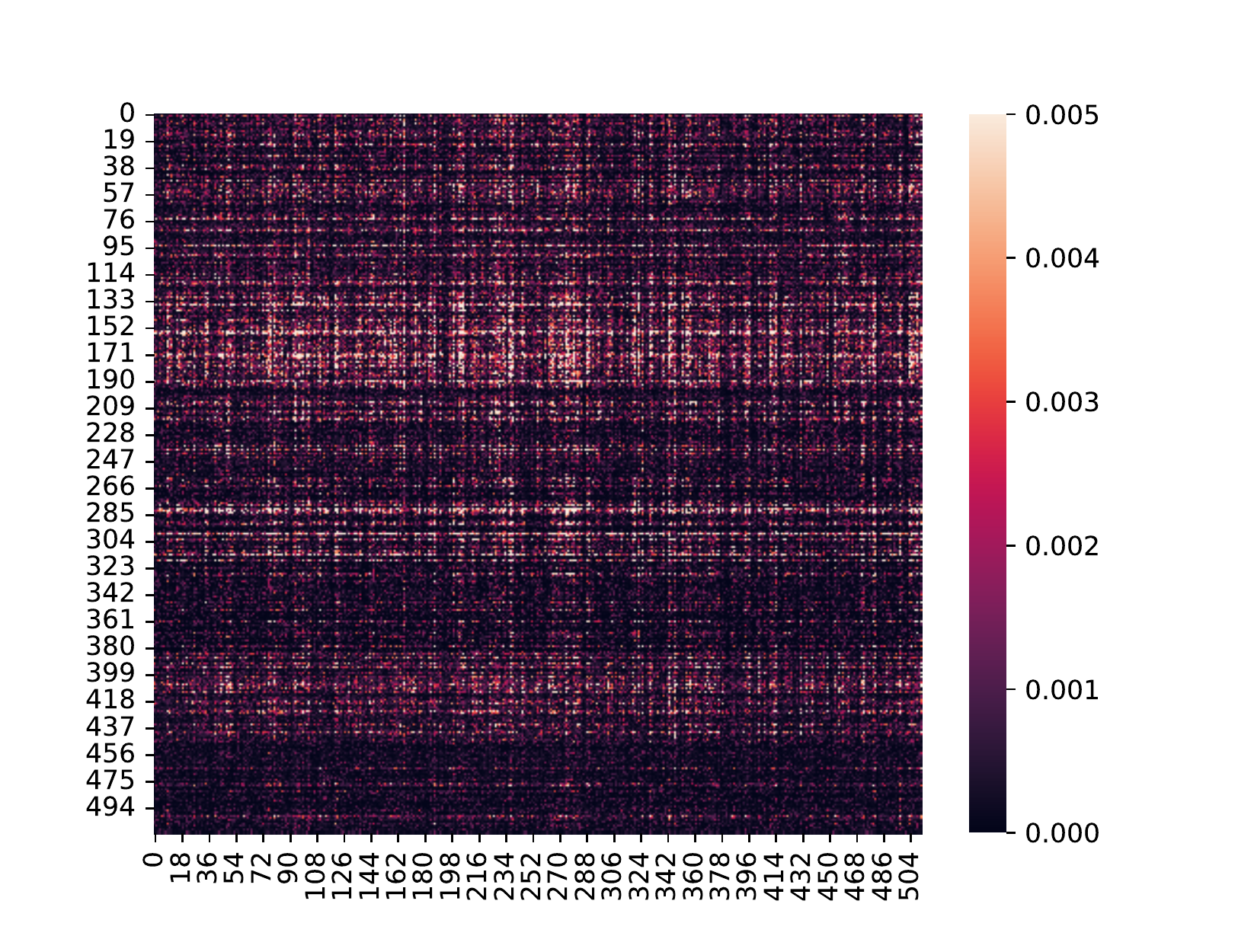}\\
	\end{tabular}
	\vspace{-0.6cm}
	\caption{\small Visualization of sparsity (black pixel) patterns in a sample query matrix $\mQ$ by GLassoformer (left) and Lassoformer (right).}
	\label{fig:sparsew}
\end{figure}




\begin{table}[!ht]
	\fontsize{9}{9}\selectfont
	\vspace{-0.3cm}
	\centering	
	\begin{tabular}{c@{\hspace{0.15cm}}c@{\hspace{0.15cm}}c@{\hspace{0.15cm}}c@{\hspace{0.15cm}}c}
		\toprule
		Models & GLasso & Lasso & Informer& 1DCNN\\
		\midrule
		Num of Params (M)     & 5.827    & 5.707    & 7.257  & 0.706  \\
		Inference Time (ms)  & 18.98    & 14.92    & 29.76  & 0.5969 \\
		\bottomrule
	\end{tabular}
	\vspace{-0.3cm}
	\caption{Comparison of model parameter size (M: million), and inference time (ms: millisecond) on GTX-1080Ti.}
	\label{tab:paramcompare}
\end{table}

\begin{table}[!ht]
	\fontsize{9.0}{9.0}\selectfont
	\vspace{-0.3cm}
	\centering
	\begin{tabular}{cccc}
		\toprule
		Models & GLasso & Informer & Lasso \\
		\midrule
		Pruning rate $ ^1 $ (\%) & 19.09 & 4.220 & 2.674\\
		Training time $ ^2 $ (second) & 9.215 & 8.975 & 8.987\\
		\bottomrule
	\end{tabular}
	\vspace{-0.3cm}
	\caption{Pruning rate and training time comparison.}
	\footnotesize{ $ ^1 $: fraction of zero query vectors (threshold = 1e-5); $ ^2 $: time per epoch}
	\label{tab:sprasetime}
\end{table}

\begin{table}[!ht]
	\fontsize{9.0}{9.0}\selectfont
	\vspace{-0.3cm}
	\centering
	\begin{tabular}{cc}
		\toprule
		 & Vanilla \\
		\midrule
		MSE(e-5) & 6.32 \\
		MAE(e-3) & 3.93 \\
		Pruning rate  (\%) & 2.3 \\
		Training time (s) & 8.75\\
		Inference Time (ms)  & 15.32 \\
		\bottomrule
	\end{tabular}
	\vspace{-0.3cm}
	\caption{Vanilla Transformer}
	\label{tab:sprasetime}
\end{table}

\vspace{-0.2cm}
\section{CONCLUSION}
\label{sec:refs}

We presented GLassoformer, a  transformer neural network model with query vector sparsity for time-series prediction, and applied it to power grid data. The model is trained through group Lasso penalty and a relaxed group-wise splitting algorithm with theoretical convergence guarantee. The model's post-fault voltage prediction is much more accurate and rapid than the recent Informer \cite{Informer_21}, also outperformed other benchmark methods in accuracy remarkably.

\newpage
{
\bibliographystyle{IEEEbib}
\bibliography{refs}
}
\end{document}